\documentclass[a4paper,conference]{IEEEtran}
\usepackage{cite}
\usepackage{hyperref}

\ifCLASSINFOpdf
  \usepackage[pdftex]{graphicx}
\else
\fi
\usepackage{xcolor}
\usepackage{tikz}
\usetikzlibrary{calc, shapes, fit}
\usepackage{pgfplots}
\pgfplotsset{compat=1.10}

\definecolor{tucol1}{rgb}{1.0 0.5 0.05}
\definecolor{tucol2}{rgb}{0.52,0.72,0.10}
\definecolor{tucol3}{rgb}{0.0 0.0 0.8}
\definecolor{tucol5}{rgb}{0.8 0.8 0}
\definecolor{tucol4}{rgb}{0.8 0 0.8}
\definecolor{tucol6}{rgb}{0 0.8 0.8}

\definecolor{tucol-kraeftig1}{RGB}{132,184,25}
\definecolor{tucol-kraeftig2}{RGB}{216,148,39}
\definecolor{tucol-kraeftig3}{RGB}{27,161,175}
\definecolor{tucol-kraeftig4}{RGB}{168,0,135}
\definecolor{tucol-kraeftig5}{RGB}{239,228,32}
\definecolor{tucol-kraeftig6}{RGB}{202,116,40}

\definecolor{tucol-kuehl1}{RGB}{132,184,25}
\definecolor{tucol-kuehl2}{RGB}{233,238,168}
\definecolor{tucol-kuehl3}{RGB}{217,233,229}
\definecolor{tucol-kuehl4}{RGB}{191,218,187}
\definecolor{tucol-kuehl5}{RGB}{113,176,96}
\definecolor{tucol-kuehl6}{RGB}{211,223,99}

\definecolor{tucol-warm1}{RGB}{132,184,25}
\definecolor{tucol-warm2}{RGB}{112,61,46}
\definecolor{tucol-warm3}{RGB}{193,178,40}
\definecolor{tucol-warm4}{RGB}{228,200,38}
\definecolor{tucol-warm5}{RGB}{178,175,132}
\definecolor{tucol-warm6}{RGB}{165,134,83}

\definecolor{tucol-rot1}{RGB}{132,184,25}
\definecolor{tucol-rot2}{RGB}{196,21,58}
\definecolor{tucol-rot3}{RGB}{157,19,42}
\definecolor{tucol-rot4}{RGB}{119,24,40}
\definecolor{tucol-rot5}{RGB}{97,126,31}
\definecolor{tucol-rot6}{RGB}{58,82,11}

\definecolor{tucol-gruen1}{RGB}{132,184,25}
\definecolor{tucol-gruen2}{RGB}{214,223,42}
\definecolor{tucol-gruen3}{RGB}{97,134,39}
\definecolor{tucol-gruen4}{RGB}{148,164,33}
\definecolor{tucol-gruen5}{RGB}{182,201,48}
\definecolor{tucol-gruen6}{RGB}{115,158,64}

\definecolor{tucol-blau1}{RGB}{132,184,25}
\definecolor{tucol-blau2}{RGB}{11,161,226}
\definecolor{tucol-blau3}{RGB}{36,123,196}
\definecolor{tucol-blau4}{RGB}{40,140,141}
\definecolor{tucol-blau5}{RGB}{177,213,230}
\definecolor{tucol-blau6}{RGB}{13,75,127}

\usepackage{siunitx}

\usepackage{amsmath, bm, amssymb}

\usepackage{tabularx}
\usepackage{multirow}
\usepackage{booktabs}
\usepackage{multirow}
\usepackage{makecell}
\newcolumntype{L}[1]{>{\hsize=#1\hsize\raggedright\arraybackslash}X}
\newcolumntype{R}[1]{>{\hsize=#1\hsize\raggedleft\arraybackslash}X}
\newcolumntype{C}[1]{>{\hsize=#1\hsize\centering\arraybackslash}X}

\begin{document}
%
\title{Self-Training of Handwritten Word Recognition for Synthetic-to-Real Adaptation}


\author{\IEEEauthorblockN{Fabian Wolf}
\IEEEauthorblockA{Department of Computer Science\\
TU Dortmund University\\
44227 Dortmund, Germany\\
fabian.wolf@cs.tu-dortmund.de}
\and
\IEEEauthorblockN{Gernot A. Fink}
\IEEEauthorblockA{Department of Computer Science\\
TU Dortmund University\\
44227 Dortmund, Germany\\
gernot.fink@cs.tu-dortmund.de}
}


%


\maketitle

\begin{abstract}
	Performances of Handwritten Text Recognition (HTR) models are largely determined by the availability of labeled and representative training samples.	
	However, in many application scenarios, labeled samples are scarce or costly to obtain.
	In this work, we propose a self-training approach to train a HTR model solely on synthetic samples and unlabeled data.
	The proposed training scheme uses an initial model trained on synthetic data to make predictions for the unlabeled target dataset.
	Starting from this initial model with rather poor performance, we show that a considerable adaptation is possible by training against the iteratively predicted pseudo-labels.
	Therefore, the investigated self-training method does not require any manually annotated training samples.
	We evaluate the proposed method on four benchmark datasets and show its effectiveness on reducing the gap to a model trained in a fully-supervised manner.
\end{abstract}

%
\IEEEpeerreviewmaketitle

\section{Introduction}
Handwritten Text Recognition (HTR) constitutes a complex pattern recognition problem, due to the high variability in its data.
A single word, even when written by the same writer, does never share the exact same appearance.
Machine learning models and especially neural networks have become the approach of choice for this recognition task.
During the last decades, the field has seen an evolution of approaches from Hidden Markov Models \cite{Bernard2011,Bluche2013,Gimenez2014} to recurrent neural networks trained with Connectionist Temporal Classification (CTC)  \cite{Graves2008,Graves2009,Wigington2017,Krishnan2018}.
Recently, sequence-to-sequence methods relying on attention based decoders got increasingly popular, pushing the performance of HTR models further on several benchmarks \cite{Bluche2017,Kang2018,Sueiras2018}.
Nonetheless, the use of increasingly complex models comes at the cost of requiring huge annotated training sets.
While datasets such as ImageNet \cite{Deng2009} contain millions of samples, handwriting data is still usually quite limited and academic benchmarks rarely surpass $100,000$ word images.
The limited number of samples is especially challenging, when confronted with document collections that have a distinctive style.
Considering, for example, historic documents, the appearance is often too specific to simply employ a HTR system trained on modern handwriting \cite{Mathew2021}.
Collection specific annotations need to be created, which often can only be done by experts making it a costly approach.
\\
While the visual appearance of different handwritten text collection may diverge heavily, the underlying structure is still well defined by character and language constraints.
This motivates the idea to pursue methods that allow to adapt a general model to a specific writing style.
Writer adaptation is a concept studied in the literature already before the field was taken over by neural networks \cite{Gilloux1994}.
A common strategy for domain adaptation in the general computer vision and object recognition communities is the use of adversarial learning, which has been also investigated to adapt a HTR model \cite{Kang2020}.
This approach relies on minimizing the differences of the domains in the feature space to improve the generalization capability of the decoder.
\\
Beside the use of unlabeled samples, a common approach to alleviate the data requirements especially in document analysis is to synthesize samples \cite{Krishnan2016, Gurjar2018, Krishnan2019, Wolf2020}.
Due to the limited variability of the synthesized data which often relies on true type fonts, performances of models purely trained on synthetic data are usually poor.
Nonetheless, bootstrapping and pre-training on synthetic samples has been shown to leverage the data problem.
\\
\begin{figure}[t]
	\centering
	\includegraphics[width=\columnwidth]{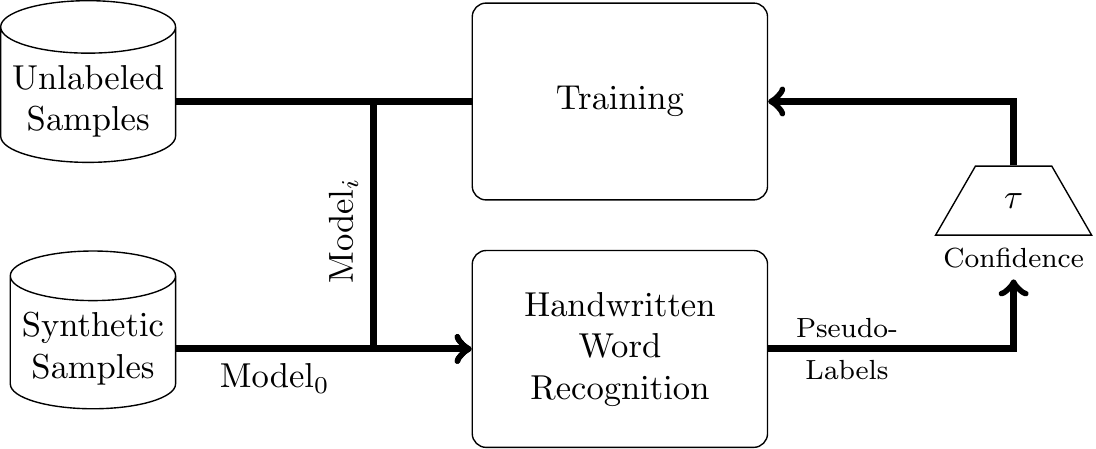}
	\caption{Self-training for Handwritten Word Recognition. First, an initial model ($\text{Model}_0$) is trained on synthetic data. Training is then performed on iteratively predicted pseudo-labels. A confidence measure is employed to threshold erroneous samples.}
	\label{fig:thumbnail}
\end{figure}
In this work, we propose a self-training method that only relies on synthetically generated handwriting samples and unlabeled in-domain data to train a handwritten word recognizer.
At the core of our approach, we use a sequence-to-sequence recognition model, that has already been investigated in the context of writer adaptation \cite{Kang2018,Kang2020}.
First, we train the model on purely synthetic samples, giving an initial model with rather poor performance.
In order to adapt from synthetic data to a document collection, the model predicts pseudo-labels for previously unlabeled samples.
Training is then continued on a selection of samples, which is based on a confidence measure. 
We show that the prediction outputs of the sequence-to-sequence model serve well as a confidence estimation despite the lack of a clear probabilistic interpretation.
Training on the iteratively predicted and selected pseudo-labels leads to high performance gains compared to the initial model.
See \autoref{fig:thumbnail} for an overview of the proposed training scheme.
We further investigate the influence of the synthetic dataset and argue that training on such encodes a form of an implicit language model.
Additionally, we include a form of consistency regularization that relies on two different augmentation approaches.

\section{Related Work}
\subsection{Handwritten Text Recognition}
Handwritten Text Recognition is a computer vision problem that has been traditionally tackled with machine learning techniques.
With text being essentially a sequence of characters represented by features, Hidden Markov Models (HMMs) were the dominant models in the early days of text recognition \cite{Bernard2011,Bluche2013,Gimenez2014}.
This is mainly due to their capability of jointly solving the recognition and segmentation task.
The rise of deep learning and neural networks was significantly influenced by handwriting recognition with MNIST being one of the first successful applications of convolutional neural networks \cite{Lecun1998}.
Recurrent architectures such as Bidirectional or Multidimensional LSTMs rapidly took over the field as they are highly effective in modeling the sequential structure of handwriting \cite{Graves2008,Graves2009,Wigington2017,Krishnan2018}.
These models have been predominantly combined with Connectionist Temporal Classification (CTC) layers to decode each single frame followed by a reduction to the resulting character sequence \cite{Graves2006}.
Due to the long term modeling capacity and the high number of parameters, it is argued that the LSTM based architectures not just learn visual information but also implicitly encode language characteristics similar to a language model \cite{Sabir2017}.
Recently, sequence-to-sequence architectures became increasingly popular.
Based on an Encoder-Decoder structure, these models are able to map from differently sized input to output sequences by the use of attention mechanisms \cite{Bluche2017,Kang2018,Sueiras2018}.


\subsection{Semi Supervised Learning (SSL)}
The widespread use of deep learning is to some extent motivated by the observation that training on larger dataset increases performances, making it often the state of the art if enough training data is available. 
Semi-supervised learning aims at exploiting the potential power of deep architecture when labeled training data is scarce or not available.
Therefore, training is based on unlabeled data in combination with a limited set of labeled samples.
%
Self-training, which is also called pseudo-labeling, has been heavily investigated in this regard \cite{Lee2013,Rosenberg2005,Berthelot2019,Berthelot2020,Sohn2020}.
Its core idea is to use labeled samples to enable the model to make predictions for unlabeled data.
The model then considers the predictions to be actual labels and training is continued respectively.
While the initial predictions are error prone, this approach is usually combined with a confidence-based selection.
A confidence measure identifies erroneous data points, which are neglected.
Additionally, works such as \cite{Berthelot2019,Berthelot2020,Sohn2020} combine self-training with consistency regularization.
By using multiple augmentation strategies, several perturbed versions of a sample are generated.
The fact that all these augmented version share the same label is then exploited for label prediction or during training by introducing a consistency loss.

Adversarial training can be considered another class of methods to adapt a model trained on one domain to a target dataset.
Commonly, an additional discriminator is added to the model supposedly to discriminate between source and target domain.
Training is performed in such a way, that the discriminator is fooled and not able to distinguish features from both domains.
This leads to learning domain independent features improving the generalization capabilities of the model \cite{Ganin2016,Tzeng2017,Peng2018}.

\subsection{SSL in Document Analysis}
Handwritten document analysis is an area of research where labeled training data is often hard to obtain.
Due to the extremely high variance in writing styles and specificities of document collections, in-domain samples are usually required to achieve state-of-the-art performances.
This makes semi-supervised approaches highly attractive and several methods have been investigated and proposed. 

In the context of document analysis, generating synthetic data is often easily possible.
A multitude of true type fonts that resemble handwriting exist and may be used to render an infinite number of synthetic training samples.
Works such as \cite{Krishnan2016,Krishnan2019,Gurjar2018} show that solely training models on these synthetic datasets yield poor performances, but it is an efficient way to pretrain and finally finetune the initial model by a limited number of target samples.
Multiple works use Generative Adversarial Networks (GANs) to generate more realistic samples of handwritten word images \cite{Alonso2019, Kang2020b, Mattick2021}.
In general, these methods either require labeled training data themselves or only marginally improve recognition performances when combining synthesized and labeled training samples.

Several works investigate adaptation strategies for optical character recognition \cite{Das2020,Kiss2021}.
Starting with models trained on related data that already achieve comparably low error rates, it is shown that performances improve with self-training.

With respect to handwriting data, self-training was shown to be effective for transductive learning in \cite{Retsinas2021}.
Other works combine self-training strategies with additional language based supervision from either a lexicon \cite{Wolf2020} or language models \cite{Tensmeyer2018}.
An approach to adapt a handwriting recognition model from synthetic to real data is proposed in \cite{Kang2020}.
By using an adversarial domain loss, the authors show that the performance gap between a model trained only on synthetic samples and the state-of-the-art performance with target data can be reduced significantly.

\section{Text Recognition}
In this work, we use a sequence-to-sequence model for HTR.
Most design choices and hyperparameters are adopted from \cite{Kang2020}.
This allows to minimize the influence of different model components and for a fair comparison to the adversarial adaptation strategy proposed in \cite{Kang2020}.
As discussed in \cite{Baek2019}, text recognition models generally follow a similar structure.
With respect to this framework the key building blocks of the HTR model can be summarized as follows.

\subsubsection{Feature Extraction}
A convolutional neural network serves as a feature extractor. 
Specifically, a VGG-19-BN architecture \cite{Simonyan2015} maps the input image $I \in \mathcal{I}$ to a two-dimensional feature map $\mathcal{X}$.
Its columns are then considered a sequence of feature vectors $(x_0, x_1,\dots,x_{N-1})$ of length $N$.
\subsubsection{Sequence Modeling}
A two layered Bidirectional Gated Recurrent Unit models the sequential nature of the input image.
Therefore, a sequence of contextual feature vectors $\mathcal{H}$ is computed.
The respective sequence of vectors $h_i \in \mathcal{H}$, $(h_0, h_1,\dots,h_{N-1})$ constitutes the final encoder output.
\subsubsection{Prediction}
The model uses an attention-based decoder to predict the final output sequence.
We use a local attention mechanism with attention smoothing to compute a mask at every time step.
Based on the attention mask, a context vector is computed, which is then fed into the decoder. 
At each time step $t$, the decoder computes a pseudo-probability distribution $d_t$ over the set of characters.
The final sequence of characters $y \in Y$, $(y_0, y_1, \dots, y_T)$ results from the character with highest pseudo-probability at each time step $t$.

The HTR model may be summarized by two functions.
First, the encoder function $\mathcal{G}_e:\mathcal{I} \rightarrow \mathcal{H}$ computes a sequence of feature vectors $h_i$.
The decoder constitutes a function $\mathcal{G}_r:\mathcal{H} \rightarrow \mathcal{Y}$ that maps the sequence of feature vectors to the final recognition sequence $y$.
During training, the recognition loss $\mathcal{L}_r$, which is essentially the cross entropy between the one hot encoded character and the pseudo-probabilities $d_t$ at every time step $t$, is minimized.
Additionally, label smoothing is applied with an $\epsilon=0.4$, regularizing the hard classification targets of 0 and 1.
For details on the model architecture and choice of hyperparameters, see \cite{Kang2018,Kang2020}.

\section{Method}
The proposed method includes two training phases.
First, an initial model is derived based on a synthetic dataset, see \autoref{sec:m_synth}.
The model is adapted to the target dataset by a self-training scheme, see \autoref{sec:m_training}, that includes consistency regularization, see \autoref{sec:m_consistency}.
\begin{figure}[t]
	\centering
	\includegraphics[width=\columnwidth]{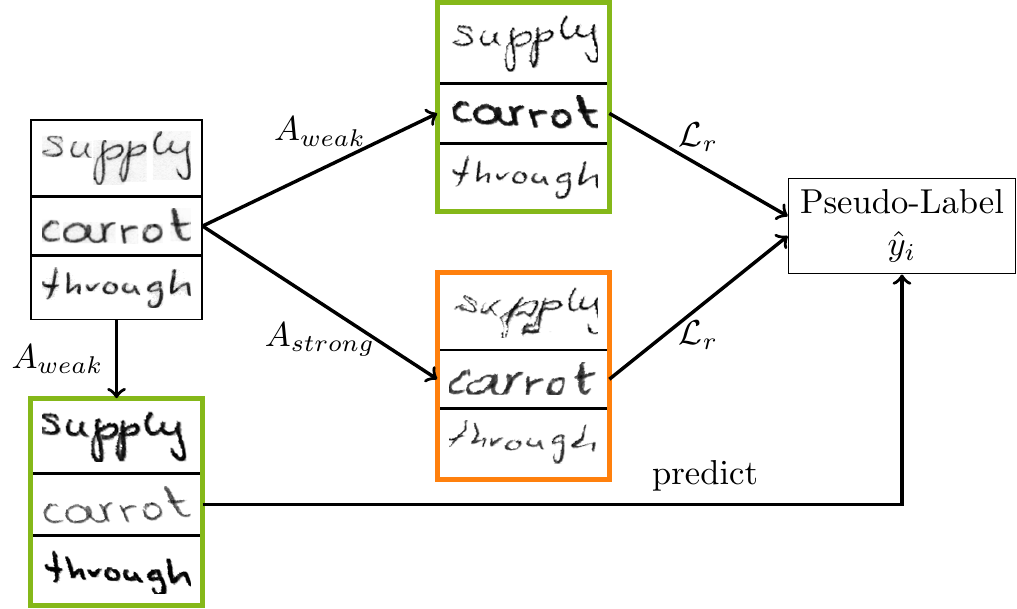}
	\caption{Consistency regularization with weak ($A_{weak}$) and strong  ($A_{strong}$) augmentations. The recognition loss $\mathcal{L}_r$ is computed between both perturbed versions of the input image and the pseudo-label $\hat{y}_i$.}
	\label{fig:consistency}
\end{figure}

\subsection{Synthetic Data Generation}
\label{sec:m_synth}
In order to train an initial model, we rely on rendering synthetic word images.
Therefore, we follow the same approach as in \cite{Krishnan2018, Wolf2020, Kang2020} and use a set of 398 true type fonts that resemble handwriting.
During synthesis, we randomly vary the stroke width, the distance between characters, the skew and slant angles.
Based on a given string, a gray scale image is rendered, again with randomly chosen intensity values for fore- and background pixels.
The resulting image is then filtered by Gaussian Smoothing.

In order to build a diverse representation of language, we collected text from the top one hundred English books of the Gutenberg Project\footnote{Text corpus based on books downloaded from \url{https://gutenberg.org}}.
The following experiments investigate the question whether the choice of text influences the model's performance or if independent character models are learned solely based on the visual appearance.
Therefore, we created three different datasets:
\begin{itemize}
	\item \textbf{Natural Text:} We rendered the individual word images directly based on the texts as occurring in the respective books.
		Character and word distributions are kept as found in the natural texts, resulting in about $13$ million word images.
	\item \textbf{Uniform:} To remove the influence of the word distribution, we determined the lexicon of unique strings of the text corpus.
		Based on a uniform word distribution, the same number of images is synthesized.
	\item \textbf{Random:} In order to further investigate the implicit language model learned form the synthetic data, we created a set of randomized word images.
		Characters are chosen randomly based on their unigram distribution.
		Word lengths follow the same distribution as found in the natural text corpora.
		All single character words and single punctuation marks that occur in the natural text are included identically.
		Furthermore, we capitalize the generated strings and append punctuation marks with the same probability as observed in the original text corpus.
\end{itemize}
\begin{figure*}[t]
	\centering
	\includegraphics[width=\textwidth]{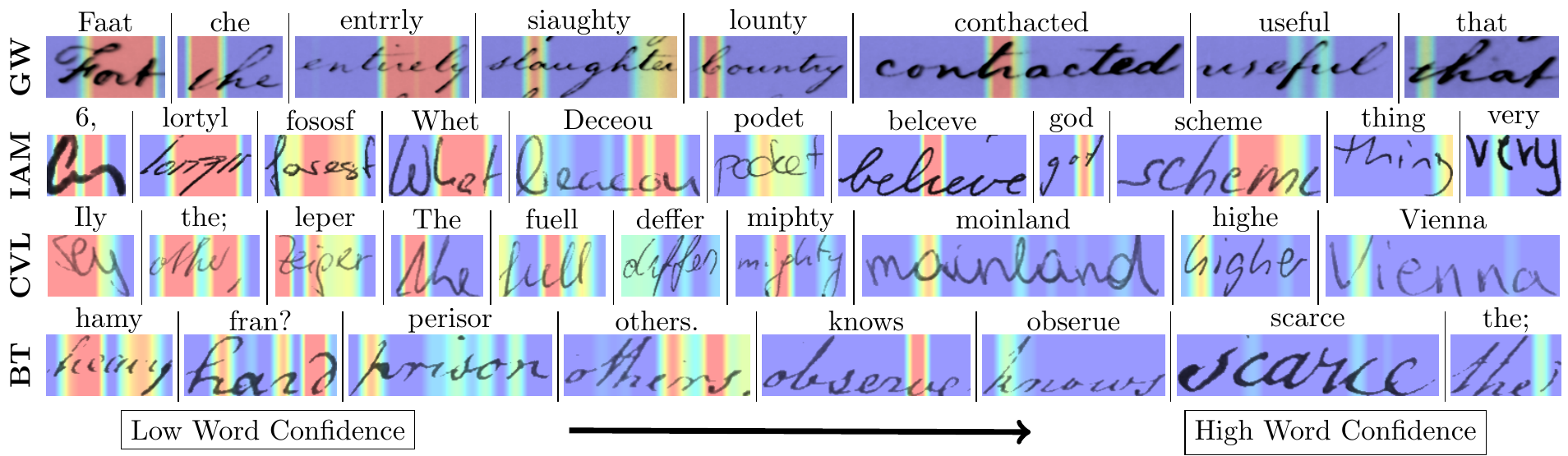}
	\caption{Exemplary predictions and confidences of the initial model. To visualize areas of lower (blue) and higher (red) confidence, we weighted the local attention masks of the model with their respective character confidences.}
	\label{fig:samples}
\end{figure*}

\subsection{Training Scheme}
\label{sec:m_training}
First, we train the recognizer for one epoch on the synthesized word images with additionally including a set of standard augmentations such as blurring or Gaussian noise and geometric transformation, i.e., shear, rotation and rescaling.
This initial model then predicts pseudo-labels $\hat{y}_i$ for the unlabeled target dataset $u_i \in \mathcal{U}, i \in \{0,1,\dots,J\}$.

As we assume that the predicted labels are highly erroneous, we aim at identifying correct samples using a confidence measure.
Based on the hidden state, the context vector and the previous embedding vector, the decoder predicts a set of pseudo-probabilities $d_t$ at each time step $t$.
For each character, we consider its corresponding activation as a numeric estimation of how confident the network is in the respective prediction.
To obtain a confidence estimation over the entire sequence, we take the mean over all character confidences:
\begin{equation}
	c = \frac{1}{T}	\sum^T_0 \max{(d_t)}
\end{equation}

After computing the word confidences over all samples in the target dataset, all samples with a confidence value below a threshold $\tau$ are neglected.
The procedure of label prediction, confidence estimation, thresholding and training is performed repeatedly for 50 cycles.

\subsection{Consistency Regularization}
\label{sec:m_consistency}
This work follows the ideas of FixMatch \cite{Sohn2020} and applies a form of consistency regularization by introducing an additional augmentation strategy.
In the following, we refer to the previously described augmentation strategy as weak augmentation.
We additionally use the grid augmentation proposed in \cite{Wigington2017} in combination with the weak augmentation.
This introduces fine grained perturbations on a character level, generally introducing stronger variations.
This allows to combine differently perturbed versions of the same sample in a single training batch, see \autoref{fig:consistency}.

In the following, $A:\mathcal{I} \rightarrow \mathcal{I}$ denotes an augmentation function.
Let $\mathcal{U} = \{u_i : i \in (0,1,\dots,B)\}$ be a batch of unlabeled samples with $c_i > \tau$ and corresponding pseudo-labels $\hat{y}_i$.
During Training each batch contains two augmented version of an image, resulting in the following formulation of the training loss $\ell$.
\begin{equation}
	\ell = \sum^B_{i=0} \mathcal{L}_r(\mathcal{G}_r(\mathcal{G}_e(A_0(u_i))), \hat{y}_i) + \mathcal{L}_r(\mathcal{G}_r(\mathcal{G}_e(A_1(u_i))), \hat{y}_i)
\end{equation}
As we compute the recognition loss with respect to a previously fixed pseudo-label, a form of consistency regularization is introduced.
The augmentation functions $A_0$ and $A_1$ constitute either a weak or strong augmentation of the image.
Implicitly, the network is not only optimized to predict a potentially erroneous pseudo-label, but also learns that both perturbed versions of the word images share the same label.

\section{Experiments}
We evaluate the proposed self-training method on four different datasets, see \autoref{ssec:datasets}.
All models are trained with ADAM optimization, a batch size of 32 and an initial learning rate of \num{2e-4}.
We train for a single epoch on the synthetic datasets.
In case of the adaptation experiments, we perform $50$ cycles of label prediction and train for one epoch on the selected portion of the entire dataset.
As a performance measure we consider character (CER) and word error rates (WER) over the respective test sets.

\subsection{Datasets}
\label{ssec:datasets}

\begin{figure*}[t]
	\centering
	\includegraphics[width=\textwidth]{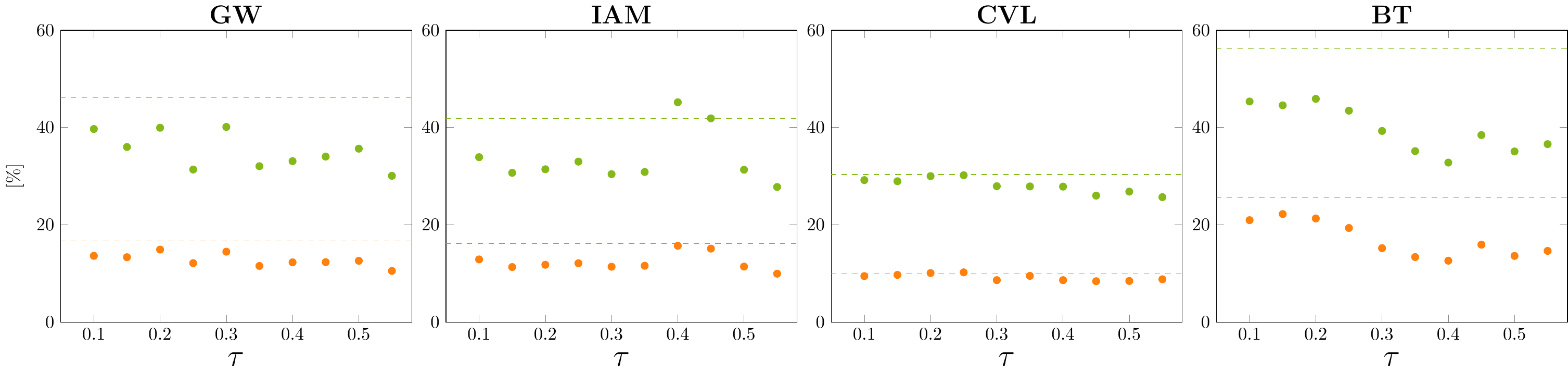}
	\caption{Performances of self-training with different thresholds for the confidence based selection. Each figure reports CERs (orange) and WERs (green). Dashed lines correspond to the performance when no selection is applied and training is performed on all samples. Training diverges for thresholds greater $0.55$, as due to the label smoothing almost no samples surpass a higher threshold.}
	\label{fig:thresh}
\end{figure*}

We present experiments on four datasets that contain handwriting in the English language.
The datasets vary considerably in their size and characteristics and include modern (IAM\cite{Marti2002}, CVL \cite{Kleber2013}) and historic documents (George Washington (GW) \cite{Lavrenko2004}, Bentham (BT) \cite{Villegas2016}).
Multi-writer datasets, such as IAM and CVL, are included to show that the proposed method is not limited to adapt to a single specific writing style.
Note, that despite the fact that the CVL dataset also includes German words, we do not make any adaptations in this regard at the synthesis stage and treat the dataset identically to the solely English corporas.
For GW, IAM and CVL, we follow the exact same evaluation protocols as described in \cite{Kang2020}.
For BT we follow \cite{Mathew2021}, as they also report recognition performances not relying on any in-domain samples.

\subsection{Synthetic Data Generation}
\label{sec:synth}

\begin{figure}[h]
	\centering
	\includegraphics[width=\columnwidth]{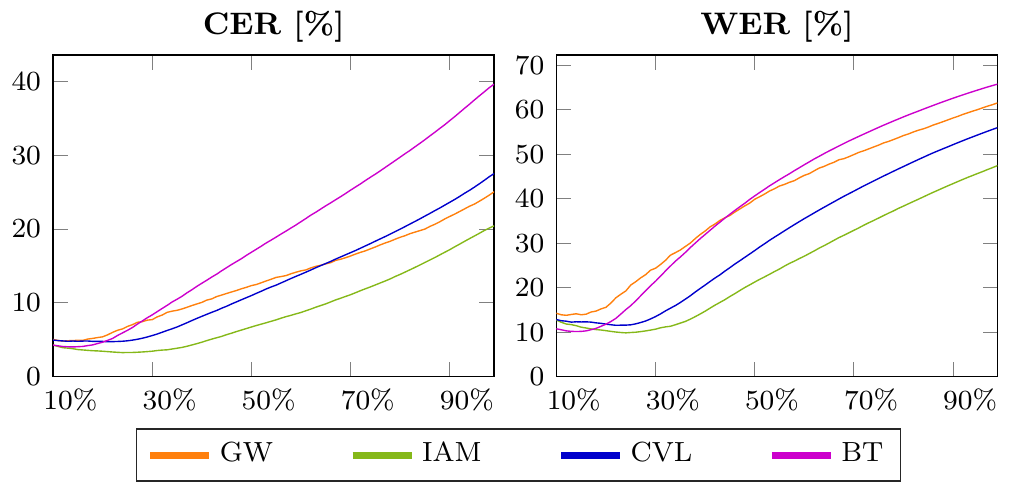}
	\caption{Error rates of the initial models on different fractions of the datasets. Samples are sorted by confidence and the error rate is calculated for the confident fraction of the dataset.}
	\label{fig:cum}
\end{figure}

\begin{table}[t]
\caption{Performances of inital models trained on synethetic datasets}
\label{tab:synth}
\scriptsize
\begin{tabularx}{\columnwidth}{L{1.3} C{0.6}C{0.6} C{0.6}C{0.6} C{0.6}C{0.6} C{0.6}C{0.6}}
\toprule
\multirow{2}{*}{Method} &\multicolumn{2}{c}{GW} & \multicolumn{2}{c}{IAM} & \multicolumn{2}{c}{CVL} & \multicolumn{2}{c}{BT}\\ [0.5ex] 
												& CER  	& WER 	 & CER 		& WER 	 & CER 	 & WER & CER 	 & WER\\
\toprule
Kang \cite{Kang2020}  & $26.05$ & $\bm{56.79}$	& $26.44$ & $54.56$ & $\bm{26.30}$ & $\bm{55.64}$ & - & - \\
\midrule
	Random 				& $40.78$ & $88.32$	& $38.13$ & $73.08$ & $35.47$ & $72.46$ & $49.88$ & $86.47$ \\
	Uniform 			& $28.97$ & $70.70$	& $34.19$ & $34.15$ & $34.46$ & $72.14$ & $43.96$ & $87.30$ \\
	Natural 			& $\bm{25.41}$ & $62.11$	& $\bm{21.78}$ & $\bm{48.07}$ & $27.51$ & $56.07$ & $\bm{39.67}$ & $\bm{65.95}$ \\
\bottomrule
\end{tabularx}
\end{table}

\begin{table}[t]
\centering
\caption{Experiments on different selection strategies}
\scriptsize
\begin{tabularx}{\columnwidth}{L{1.7} R{0.6}L{0.6} R{0.6}L{0.6} R{0.6}L{0.6} R{0.6}L{0.6}}
\toprule
\multirow{2}{*}{Selection} &\multicolumn{2}{c}{GW} & \multicolumn{2}{c}{IAM} & \multicolumn{2}{c}{CVL} & \multicolumn{2}{c}{BT}\\ [0.5ex] 
												& CER  	& WER 	 & CER 		& WER 	 & CER 	 & WER & CER 	 & WER \\
\toprule
	Initial 			& $25.41$ & $62.11$	& $21.78$ & $48.07$ & $27.51$ & $56.07$ & $39.67$ & $65.95$ \\
\midrule
	None 									& $16.65$ & $46.13$	& $16.19$ & $41.89$ & $10.21$ & $30.91$ & $25.54$ & $56.16$ \\
	Thresholding				  & $12.37$ & $32.65$	& $\bm{9.96}$ & $\bm{27.78}$ & $ 8.76$ & $25.69$ & $\bm{14.64}$ & $\bm{36.58}$ \\
	Random$^{(*)}$ 					& $16.75$ & $43.47$ & $13.99$ & $37.60$ & $ 8.89$ & $27.19$ & $16.66$ & $50.52$ \\
	Confidence$^{(*)}$ 			 		& $\bm{12.28}$ & $\bm{31.62}$ & $10.95$ & $29.07$ & $ \bm{8.62}$ & $\bm{23.31}$ & $14.72$ & $37.06$ \\
\bottomrule
\end{tabularx}
	\raggedleft ($*$) predefined schedule for fixed portions
	\label{tab:conf}
\end{table}

\begin{table}[t]
\centering
\caption{Consistency Regularization}
\scriptsize
\begin{tabularx}{\columnwidth}{C{1} C{1} C{0.6}C{0.6} C{0.6}C{0.6} C{0.6}C{0.6} C{0.6}C{0.6}}
\toprule
\multirow{2}{*}{$A_0(\cdot)$} & \multirow{2}{*}{$A_1(\cdot)$} &\multicolumn{2}{c}{GW} & \multicolumn{2}{c}{IAM} & \multicolumn{2}{c}{CVL} & \multicolumn{2}{c}{BT}\\ [0.5ex] 
					&		& CER  	& WER 	 & CER 		& WER 	 & CER 	 & WER & CER 	 & WER \\
\toprule
	weak 		& None				& $12.37$ & $32.65$	& $ 9.96$ & $27.78$ & $ 8.76$ & $25.69$ & $14.64$ & $36.58$ \\
	strong 	& None				& $16.17$ & $44.24$	& $ 9.87$ & $26.44$ & $ 7.84$ & $24.71$ & $15.31$ & $35.81$ \\
	weak 		& weak 		& $11.74$ & $32.99$	& $11.04$ & $30.98$ & $ \bm{7.59}$ & $\bm{21.34}$ & $11.39$ & $29.12$ \\
	weak    & strong 	& $\bm{10.53}$ & $\bm{30.07}$	& $ \bm{8.76}$ & $\bm{25.26}$ & $ 8.21$ & $25.69$ & $ \bm{9.59}$ & $\bm{26.35}$ \\
\bottomrule
\end{tabularx}
\label{tab:cons}
\end{table}

We train on synthetically generated word images to derive an initial model.
\autoref{tab:synth} presents resulting error rates based on the three different generation procedures.
Training on the dataset derived from the natural text corpora yields similar performances to \cite{Kang2020}.
Despite minor differences in the synthesis pipeline, the results underline that models with similar performances are derived.
Therefore, a fair comparison of the adaptation strategies is possible, avoiding influences of the synthetic data quality.
Removing the natural word distribution as well as training on randomly generated text strings results in a drop of performance.
This indicates that the language characteristica encoded in the synthetic data generation process contribute to the resulting performance.

\begin{table*}
	\centering
	\caption{Comparison to the State of the art}
	\begin{tabularx}{\textwidth}{C{3.0} L{2.0} R{0.6}L{0.6} R{0.6}L{0.6} R{0.6}L{0.6} R{0.6}L{0.6}}
\toprule
& \multirow{2}{*}{Method} & \multicolumn{2}{c}{GW} & \multicolumn{2}{c}{IAM} & \multicolumn{2}{c}{CVL} & \multicolumn{2}{c}{BT}\\ [0.5ex] 
&													& CER  	& WER 	 & CER 		& WER 	 & CER 	 & WER & CER 	 & WER\\
\toprule
\multirow{1}{*}{ Real target only} 
& Kang et al. \cite{Kang2020}	& $ 4.56$ & $13.49$	& $ 6.88$ & $17.45$ & $ 3.64$ & $ 7.77$ & - 			& - \\
\midrule
\multirow{3}{*}{Synthetic only}
& Kang et al. \cite{Kang2020}	& $26.05$ & $56.79$	& $26.44$ & $54.56$ & $26.30$ & $55.64$ & - 			& - \\
&	Mathew et al. \cite{Mathew2021}		& - 			& - 			& - 			& - 			& - 			& - 			& - 			& $86.28$ \\ 
&	Ours 							& $25.41$ & $62.11$	& $21.78$ & $48.07$ & $27.51$ & $56.07$ & $39.67$ & $65.95$ \\
\midrule
\multirow{2}{*}{Unsupervised adaptation} 
&	Kang et al. \cite{Kang2020} & $16.28$ & $39.95$	& $14.05$ & $34.86$ & $19.19$ & $44.29$ & - & - \\
& Ours  						& $10.53$ & $30.07$	& $ 8.76$ & $25.26$ & $ 8.21$ & $25.69$ & $ 9.59$ & $26.35$ \\
\bottomrule
\end{tabularx}

	\label{tab:sota}
\end{table*}

\subsection{Confidence Estimation}
After training an initial model on the synthetic dataset, we investigate the adaptation capability of the self-training approach. 
As shown in \autoref{tab:conf}, training on all pseudo-labeled samples with no confidence based selection already leads to performance gains compared to the initial model.

The first question to raise is whether the output activations are a suitable measure to identify erroneous samples.
See \autoref{fig:samples}, for exemplary prediction results of the initial model solely trained on synthetic data.
Furthermore, we weighted the attention masks of the model with the respective character confidence in order to visualize the area of high and low confidence.
Qualitatively, it can be observed that characters or parts of words with low confidences coincide with recognition errors.
Image areas of low confidence are often ambiguous and generally result in plausible errors.
To further underline the expressiveness of the proposed confidence measure, we show error rates over differently confident parts of the datasets in \autoref{fig:cum}.
Lower error rates can be observed in the more confident parts of the dataset, leading to the conclusion that the output activation effectively results in a numeric value of how confident the network is in its prediction.

In a first experiment, we investigate the influence of the confidence measure.
See \autoref{fig:thresh}, for an overview on the performance gains achieved with different threshold values on all datasets.
Thresholding the confidence measure and only performing training on more confident parts of the dataset, consistently improve performances compared to training on all samples.
Especially, applying a high confidence threshold of $\tau=0.55$ results in performance gains over all datasets.
Note, that tuning the threshold is generally not possible in the application scenario, as a validation set requires labeled sample.
Nonetheless, we appreciate that the threshold value seems to be quite independent from the considered dataset and that, in general, thresholding improves performances.
\autoref{tab:conf} presents the error rates when a confidence threshold of $\tau=0.55$ is incorporated in the self-training scheme.

During training, we observe that an increasing number of samples lie above the threshold.
To further underline the expressive power of the confidence measure, we conducted the following experiment.
During the first $10$ cycles of pseudo-labeling, we train on the most confident $60\%$ of the dataset, followed by training on $80\%$ and $100\%$ for another $20$ cycles each, respectively.
Using predefined portions of the dataset allows to remove the influence of the confidence measure on the number of selected samples.
Therefore, it is possible to directly compare a confidence based selection to simply taking random portions of the dataset.
While randomly selecting samples achieves similar performances to training on all samples, the proposed confidence measure improves results.
Nonetheless, thresholding the confidence measure performs better in most cases and does not introduce any further hyperparameters except the actual threshold value.

\subsection{Consistency Regularization}
As shown in \autoref{tab:cons}, introducing consistency regularization leads to performance gains.
We report results for extending the training batch by either an additional weakly or strongly augmented version of the training images.
In order to account for the higher number of seen samples, we increased the number of self-training cycles to $100$ which did not lead to performance gains.
Furthermore, we report results for exclusively training on strongly augmented samples.
In our experiments, we observe performance gains already when adding a weakly augmented image to the training batch, indicating the regularization power of the consistency approach.
Combining strongly and weakly augmented versions and enforcing consistency between them gives the highest performances in our experiments.
This is consistent with the results reported for the evolution from MixMatch to FixMatch as investigated in \cite{Berthelot2019,Berthelot2020,Sohn2020}.

\subsection{Comparison}
\autoref{tab:sota} compares our results to the literature.
To this end, we use a threshold of $\tau=0.55$ and apply consistency regularization with weak and strong augmentations.
The work of Kang et al. proposes to perform adaptation from synthetic to real data based on an adversarial domain loss \cite{Kang2020}.
The proposed pseudo-labeling approach significantly outperforms the domain loss strategy proposed in \cite{Kang2020}.
We argue that the performance differences are mainly due to the different adaptation methods, as we avoided any model differences and report similar performances before adaptation.
A drawback of the domain loss strategy proposed in \cite{Kang2020} is that it only aims at regularizing the feature space in such a way that synthetic and real samples are indistinguishable.
The observed performance gains are then the result of an implicitly improved generalization capability.
In contrast, a pseudo-labeling strategy is highly feasible in the handwriting domain, due to heavy constraints on the possible recognition results. 
A significant portion of the performance gains may be explained by the implicit language model learned from synthetic data which is then exploited during label prediction.
Interestingly, the approach also leads to considerable performance gains even when the initial model is comparably poor, as shown in the experiments on BT.
This further encourages the consideration of adaptation strategies as considerable improvements can be achieved when compared to models solely trained on related data, as presented in \cite{Mathew2021}.

\section{Conclusions}
In this work, we propose a self-training method for adapting a sequence-to-sequence HTR model from synthetic to real data.
We show that it is possible to use the output activations as confidence estimations, which may be exploited to identify more accurate portions of the pseudo-labeled dataset.
Adding consistency regularization techniques leads to further performance gains.
The proposed approach is superior to the previously investigated adversarial strategy and we report the highest performances on four dataset with respect to training only on synthetic and unlabeled data.
Due to the constraints present in language, we encourage the investigation of self-training approaches in handwriting recognition and document analysis in general as a potentially powerful possibility to reduce the data demand of deep learning models.


\clearpage



\bibliographystyle{IEEEtran}
\bibliography{literature}
%
%
%

\end{document}